\documentclass{article}

\usepackage[final,nonatbib]{neurips_2022}
\usepackage[utf8]{inputenc} 
\usepackage[T1]{fontenc}    
\usepackage{hyperref}       
\usepackage{url}            
\usepackage{booktabs}       
\usepackage{amsfonts}       
\usepackage{nicefrac}       
\usepackage{microtype}      
\usepackage{xcolor}

\usepackage{pdfpages}
\usepackage{url}
\usepackage{booktabs}
\usepackage{hyperref}
\usepackage[utf8]{inputenc}
\usepackage[pangram]{blindtext}
\usepackage{lipsum}
\usepackage{fancyhdr}
\usepackage{amsmath}
\usepackage{multirow}
\usepackage{diagbox}
\usepackage{float}
\usepackage{algorithm}
\usepackage{soul}
\usepackage{algpseudocode}

\title{On Adversarial Examples for Text Classification\\by Perturbing Latent Representations}

\author{%
  Korn Sooksatra \quad Bikram Khanal \quad Pablo Rivas\thanks{Authors performed this work while at the \href{www.baylor.ai}{Baylor.AI} lab.} \\
  School of Engineering \& Computer Science \\
  Department of Computer Science\\
  Baylor University, Texas, USA \\
  \texttt{\{Korn\_Sooksatra1,Bikram\_Khanal1,Pablo\_Rivas\}@Baylor.edu}\\}

\begin{document}
\maketitle
\begin{abstract} \label{sec:abstract}
Recently, with the advancement of deep learning, several applications in text classification have advanced significantly. However, this improvement comes with a cost because deep learning is vulnerable to adversarial examples. This weakness indicates that deep learning is not very robust. Fortunately, the input of a text classifier is discrete. Hence, it can prevent the classifier from state-of-the-art attacks. Nonetheless, previous works have generated black-box attacks that successfully manipulate the discrete values of the input to find adversarial examples. Therefore, instead of changing the discrete values, we transform the input into its embedding vector containing real values to perform the state-of-the-art white-box attacks. Then, we convert the perturbed embedding vector back into a text and name it an adversarial example. In summary, we create a framework that measures the robustness of a text classifier by using the gradients of the classifier.
\end{abstract}
  
\section{Introduction} \label{sec:intro}



Over the decades of research in Deep Learning, it has contributed to improving numerous machine learning applications. While there are several perks of deep learning, Goodfellow \emph{et al.} \cite{goodfellow2014explaining} showed that machine learning was vulnerable to adversarial examples. An adversary has access to different examples that he can use to mislead his target, a model. Unlike machine-generated models, humans are not affected by those adversarial attacks. An adversary generates adversarial examples by computing tiny imperceptible perturbations and adding them to unadulterated samples. Several works \cite{madry2017towards, papernot2016limitations, sooksatra2021enhancing,sooksatra2022evaluation,sooksatra2021evaluating,su2019one} attempted to utilize adversarial examples to evaluate the robustness of a learning model.

Nonetheless, many of these works focus on adversarial examples in computer vision. These examples are effortless to generate due to the availability of many pixels to perturb and the fact that perturbations can take on almost any real value.
However, the literature only provides a few works focusing on producing adversarial examples in Natural Language Processing (NLP) due to the discrete input. Furthermore, almost all the works in NLP have directly changed the discrete values of input (e.g., changing words, changing characters, adding words, and removing words) by designing black-box attacks. These attacks are given only the confidence of text classifiers. Nonetheless, these approaches may need to be revised to create fluent and natural texts.

In this work, we create adversarial examples from the embedding vectors of inputs to apply the state-of-the-art white-box adversarial attacks widely used in the field of computer vision. These attacks are given all the information of text classifiers. Our approach is based on the work in \cite{zhao2017generating} that designed a black-box attack to find adversarial examples on the embedding vectors of images and texts. However, the work focused little on creating adversarial examples for texts. Therefore, our contributions are summarized as follows:
\begin{itemize}
    \item We implement the encoder and decoder and their training scheme that can generate embedding vectors for a specific task.
    \item Our approach is among the first that applies a white-box adversarial attack on the embedding vectors of texts to generate adversarial examples. 
    \item We extensively construct experiments showing that our approach can produce natural adversarial examples.
\end{itemize}

The rest of this paper is organized as follows: Section \ref{sec:related_works} briefly discusses the state of the art, while Section \ref{sec:probformulation} specifies the problem we address. Our approach to the problem and the results obtained are discussed in Section \ref{sec:approach} and \ref{sec:results}, respectively. Finally, conclusions are drawn in Section \ref{sec:conclusions}.

\section{Related Works \label{sec:related_works}}
Some existing works also desired to find adversarial examples for applications in the field of NLP. Jia and Liang \cite{jia2017adversarial} created a method to mislead a reading comprehension system by adding an adversarial sentence to a paragraph. Iyyer \emph{et al.} \cite{iyyer2018adversarial} proposed a framework to generate an adversarial example of a clean text by finding its paraphrase such that this adversarial example's semantic could be the same as the clean text. Alzantot \emph{et al.} \cite{alzantot2018generating} applied an evolutionary algorithm to select a word in a text to change to its close word to generate an adversarial example. Ebrahimi \emph{et al.} \cite{ebrahimi2017hotflip} constructed a white-box attack by utilizing a targeted model's gradient to flip some characters in a text to create an adversarial example. Jin \emph{et al.} \cite{jin2020bert} showed that BERT \cite{devlin2018bert} was not very robust by creating an attack that utilized the confidences of a targeted model to replace some words in a text with their close-semantic words to generate an adversarial example. Zang \emph{et al.} \cite{zang2020learning} applied the policy gradient to find words in a text to replace with other words to create an adversarial example by considering the history information.

To the best of our knowledge, no work has constructed a white-box attack to perturb the latent representation of a text to create an adversarial example. 

\section{Problem Formulation} \label{sec:probformulation}
Given a text classifier $C$ (e.g., a sentiment analyzer and a news-type classifier), our goal is to evaluate the $C$ by finding misclassified samples (adversarial examples). We compute small perturbations $\delta$ and add them to an input of $C$ such that the prediction is not the same as its ground-truth class. That is, given an input $x$ and its ground-truth class $y$, we compute $\delta$ such that $C(x) = y$ and $C(x^{*}) \neq y$ where $C$ predicts the class of $x$ and $x^{*} = x + \delta$. However, the inputs are discrete in $C$; hence any changes to the input are obvious. Therefore, we find an embedding vector of $x$ and compute $\delta$ instead. Then, we transform the perturbed embedding vector back to text. The next section will explain the mechanism to transform a text to an embedding vector and vice versa to guarantee that the text and its reconstruction belong to the same semantic.

\section{Our Approach}
\label{sec:approach}

\begin{figure}[!ht]
    \centering
    \includegraphics[width=0.9\textwidth]{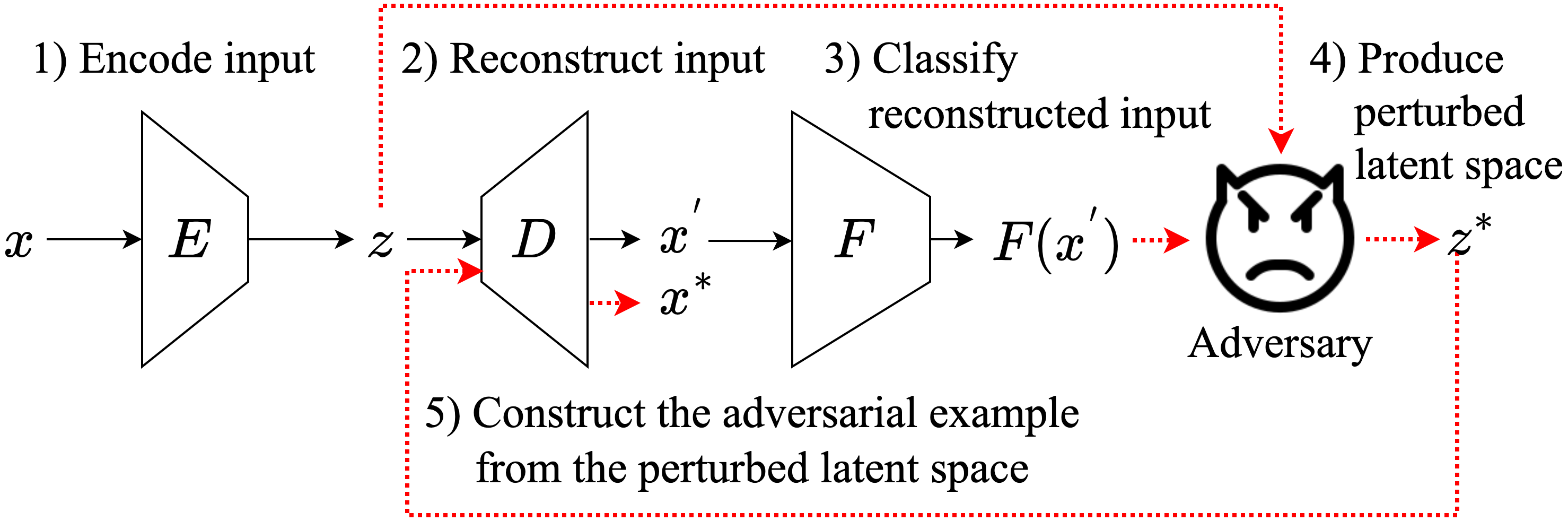}
    \caption{The framework and flow of our approach. It include the encoder ($E$), the decoder ($D$), the targeted classifier ($F$) and the adversary who performs an adversarial attack.}
    \label{fig:architecture}
\end{figure}

Our framework consists of three main components: the encoder ($E$), the decoder ($D$), and the adversary. Those encoder and decoder are neural-network-based, and the adversary can be any adversarial attack (e.g., FGSM \cite{goodfellow2014explaining} and PGD \cite{madry2017towards}). The framework works as follows: a) Given an input $x$ (i.e., a set of tokens), we feed it through the encoder to obtain its embedding vector $z$. b) $z$ is then fed to the decoder to obtain the regenerated text $x^{'}$ (i.e., a set of tokens) whose semantic does not change from $x$. Note that the initial input is the start token, and its output is the input of the next time step. This mechanism happens until the output is the end-of-sequence token. c) We pass $x^{'}$ to the targeted classifier and obtain the gradients $\nabla_z l(F(x^{'}), y_x)$ where $l(F(x^{'}), y_x)$ is a loss function between the classifier's confidence-score vector on $x^{'}$ (i.e., $F(x^{'})$) and the ground-truth vector of $x$'s label (i.e., $y_x)$. d) We apply an adversarial attack to obtain an adversarial example $z^{*}$ of $z$ by using the obtained gradients and the embedding vector $z$. e) Then, we generate a text by passing $z^{*}$ through the decoder and obtain an adversarial example $x^{*}$ of $x$. Figure \ref{fig:architecture} illustrates our framework and how the information flow in our framework, and Algorithm \ref{algo:adversarial_examples} explains the framework

\begin{algorithm}[h!]
\caption{Generating adversarial examples.}
\label{alg:cap}
\textbf{Input}: $x, E, D, F$\\
\textbf{Output}: Adversarial example: $x^{*}$
\begin{algorithmic}[1] 
\State $z \gets E(x)$ \Comment{Encode $x$}
\State $x^{'} \gets D(z)$ \Comment{Reconstruct $x$ from $z$}
\State $z^{*} \gets Attack(F(x^{'}), z)$ \Comment{Note that $Attack$ can be one of gradient-based adversarial attacks}
\State $x^{*} \gets D(z^{*})$ \Comment{Create an adversarial example from $z^{*}$}
\end{algorithmic}
\label{algo:adversarial_examples}
\end{algorithm}

\subsection{Encoder and Decoder}
Noticeably, our approach strongly depends on the encoder and decoder. Thus, we need to efficiently train them such that the distance between two embedding vectors can indicate how much the semantics of their particular text are different. Then, we train the encoder and decoder together, and the reconstruction loss is the average categorical cross-entropy between the input and its reconstruction. However, only this training cannot ensure that the embedding vector represents the semantics of the input for the targeted classifier. Therefore, we also create a classifier whose input is an embedding vector and output is the same as the targeted one. Then, we train it together with the encoder and decoder. With the classifier, we can well organize the embedding vectors. Also, this training prevents the encoder from overfitting the reconstruction task, where it only outputs the same embedding vector regardless of the input. Then, the loss function of this training is
\begin{equation*}
    R(X, X^{'}) + \lambda L(c(E(X)), Y_X),
\end{equation*}
where $X$ is a batch of input, $Y_X$ is the labels of $X$, $R(X, X^{'})$ is the reconstruction loss between $X$ and its reconstruction $X^{'}$ by measuring the average categorical cross-entropy loss of the tokens in all positions, $E(X)$ is the encoder function of $X$, $c(E(X))$ is our classifier whose input is an embedding vector, $L(c(E(X)), Y_X)$ is the categorical cross-entropy loss between the predicted vector from the classifier by given the embedding vector of $X$ (i.e., $c(E(X))$) and the ground-truth vector of $X$ (i.e., $Y_X$) and $\lambda$ is the balancer between the two loss functions. Algorithm \ref{algo:encoder_decoder} demonstrates how we train the encoder and decoder. After obtaining the encoder and decoder, we can perform our framework to find adversarial examples with the earlier process.

\begin{algorithm}[h!]
\caption{Training encoder and decoder.}
\label{alg:ted}
\textbf{Input}: $X, Y_X, E, D, c$\\
\textbf{Output}: Trained $E$, trained $D$
\begin{algorithmic}[1] 
\While{ $E, D, c$ are not converge}

\State $X^{'} \gets D(E(X))$ \Comment{Obtain reconstruction of $X$}
\State Train $E, D, c$ with respect the loss function $  R(X, X^{'}) + \lambda L(c(E(X)), Y_X)$
\EndWhile
\end{algorithmic}
\label{algo:encoder_decoder}
\end{algorithm}

\subsection{Adversarial Attacks}
The adversarial attacks in this work focus on perturbing the embedding vectors of inputs instead of the inputs themselves due to the discrete values of the inputs. We choose Fast Gradient Sign Method (FGSM)~\cite{goodfellow2014explaining} as the attack because it is fast and enough to show that our approach is effective.

\subsection{Fast Gradient Sign Method (FGSM)}
When the adversary obtains the embedding vector $z$ of input $x$ and the gradients $\nabla_z l(F(x^{'}), y_x)$, it attempts to find perturbation such that the perturbed embedding vector can generate a reconstructed set of tokens that mislead the targeted classifier. That is, this attack creates an adversarial example by
\begin{equation*}
    z^{*} = z + \epsilon \cdot \text{Sign}(\nabla_z l(F(x^{'}), y_x)),
\end{equation*}
where $\epsilon \cdot \text{Sign}(\nabla_z l(F(x^{'}), y_x))$ is the computed perturbation, $\epsilon$ is the perturbation bound and $\text{Sign}(v)$ outputs only the signs of $v$. Explicitly, this attack performs only one step. 


\section{Experiments and Results}
\label{sec:results}

This section explains our experiments and discusses the results. We applied our framework, and the results were seemingly natural, as if they were not under attack.

\subsection{Experimental Setup}
First, we describe the targeted classification task and how we train the targeted classifier for the particular task. Further, we explain the training of the encoder and decoder of the framework and adversarial attack setups. Unfortunately, we do not have any relevant work that can compare our approach to. We use the BERT preprocessor to convert a plain text to a set of 30522 different tokens.

Furthermore, we used the Ag News corpus \cite{zhang2015characterlevel} for our targeted task. This corpus is a collection of the news; given a title and description, the task is to predict the news topic. There are four topics: World \textbf{(class W)}, Sports \textbf{(class S)}, Business \textbf{(class B)} and Sci/Tech \textbf{(class S/T)}. Each topic has $30000$ samples in the training corpus, and the test corpus has $7600$. We create a classifier using the BERT preprocessor to convert a text to a set of tokens. The following layers in our architecture are listed as follows: an embedding layer whose embedding vector's size is 512, 2 layers of LSTMs \cite{hochreiter1997long} with 128 and 64 neurons, respectively, and a dense layer with four neurons with the Softmax activation as the output layer. We trained the classifier with the Adam optimizer \cite{kingma2014adam} and batch size of 256 in 5 epochs and achieved a validation accuracy on the test corpus of $90\%$.

For the encoder and decoder, we choose a pre-trained BERT with the embedding vector size of 512 as our encoder since it is one of the state-of-the-art text encoders. Also, we created our decoder consisting of an embedding layer of 512 embedding vector size, two layers of LSTMs \cite{hochreiter1997long} with 512 neurons, and a dense layer with 30522 neurons with the Softmax layer as the output layer. Furthermore, we create a classifier for the Ag News corpus with only one dense layer of 4 neurons with the Softmax activation function as the output layer. Note that this classifier receives an embedding vector from the pre-trained BERT as its input. These three parts (i.e., the encoder, decoder, and classifier) are all trainable, and we jointly train them with the Adam optimizer \cite{kingma2014adam} and batch size of 256. Table \ref{tab:ag_news_samples} shows some examples of texts and their corresponding reconstruction. Although the reconstructed texts are different from their original ones, they are of the same type. For example, the texts in the first row are in the type of sports news, and the ones in the second row are in the type of world news.

\begin{table}[!ht]
    \centering
    \caption{Examples of texts in the test Ag News corpus and their particular reconstruction generated by the encoder and decoder.}
    \begin{tabular}{|p{.96\textwidth}|}
    \hline\\
        \textbf{Original text (Class S): } \texttt{us cyclists capture three medals athens , greece - tyler hamilton \# 39 ; s greatest ride capped the finest olympic day for us cycling , which won three of the six medals awarded in wednesday \# 39 ; s road time trials - surpassing its two total road medals won since the 1984 games in los . . .}  \\\\
        \textbf{Reconstruction (Class S): } \texttt{cricket : aussies crowing but india \# 39 ; s grip on stump india \# 39 ; s cricket board praiseds shane warne on monday as the first test against australia captain nagpur was the buttreded his team \# 39 ; s chances for a test against australia .}\\\\
        \hline\\
        \textbf{Original text (Class W): } \texttt{burundi urges sanctions on rebels over massacre dar es salaam ( reuters ) - burundi will urge regional leaders at a summit on wednesday to impose sanctions against the rebel hutu forces for national liberation ( fnl ) , blamed for the slaughter of more than 160 congolese tutsi refugees in . . .}\\\\
        \textbf{Reconstruction (Class W): } \texttt{israeli troops kill two palestinians in gaza ( reuters ) reuters - israeli troops shot dead two palestinian gunmen on wednesday in a clash with israeli troops and militants in northern gaza , witnesses said .}\\ \\
        \hline
    \end{tabular}
    \label{tab:ag_news_samples}
\end{table}

\begin{table}
    \centering
    \caption{Examples of texts in the test Ag News corpus and their particular reconstruction generated by the encoder and decoder.}
    \begin{tabular}{|p{.96\textwidth}|}
    \hline\\
        \textbf{(Class S/T $\longrightarrow$ W)} \texttt{u . s . to share funds for more ( ap ) ap - the nation ' s top education department is planning to raise a new government research program in 2005 and plans to begin issuing new and \st{negative effects on the scale of the nation ' s biggest cities .}{\color{red} more popular voting machines in the united states .}}\\\\
        \hline\\
        \textbf{(Class S $\longrightarrow$ W)} \texttt{astros beat rockies to win nl playoff spot houston ( reuters ) - the houston astros have \st{picked up their first playoff berth in five years , their first big one - day winning streak in a season - clinching victory , the houston astros made the playoffs finale for their 13th straight year .} {\color{red} found a huge win over the houston astros with a huge win on their national league championship series at the houston astros .}}\\\\
        \hline\\
        \textbf{(Class B $\longrightarrow$ S/T)} \texttt{google shares surge in debut on market share shares of google , the internet search engine , said its first - half profit rose 39 percent \st{, boosted by strong results in its international business .} {\color{red} as it priced its online rental market .}}\\\\
        \hline
    \end{tabular}
    \label{tab:ag_news_adversarial_examples}
\end{table}

\subsection{Results}
In this section, we evaluate our targeted model with respect to the robustness by using FGSM. We consider the attack to be successful on text $x$ when: a) the reconstructed text generated by the encoder and decoder is classified as the same type as $x$ by the targeted model. b) the adversarial example generated by the decoder and the embedding vector perturbed by the attack is classified as a different type from $x$ by the targeted model.

We used Ag News corpus for this evaluation and the perturbation bound ($\epsilon$) of $0.05$.
Table \ref{tab:ag_news_adversarial_examples} shows some results of our approach including the clean texts and their corresponding adversarial examples. Noticeably, the adversarial examples are classified as different classes from their clean texts although only a few words at the end changed. Furthermore, the content in the adversarial examples and their predicted classes are also explicitly counter-intuitive to humans. 

Additionally, Table \ref{tab:percentage_fgsm_agnews} demonstrates the interesting result because the Sci/Tech-news samples are the most vulnerable. Explicitly, the Sci/Tech-news samples are very easy to transform to World news or Business news. However, it is not very easy to transform World news to Sci/Tech news. Moreover, the Sport-news samples are also trivial to transform to World news as shown in the table. Nonetheless, it is not easy to transform World news to Sport news. Therefore, World news is very difficult to find adversarial examples, but many adversarial examples of other news belong to World news. This phenomenon is interesting.

\begin{table}[!ht]
    \centering
    \caption{The percentage of the successfully generated adversarial examples in each source class in Ag News corpus. The rows are the source texts' predicted classes, and the columns are the predicted classes of their corresponding adversarial examples.}
    \begin{tabular}{c|cccc}
    \toprule
     \diagbox[width=10em]{Clean}{Adv}    &  World  & Sport & Business & Sci/Tech\\
    \midrule
        World & - & 2.4\% & 6.4\% & 6.4\%\\
        Sport & 27.7\% & - & 0\% & 0.7\%\\
        Business & 9.8\% & 0.7\% & - & 14.9\%\\
        Sci/Tech & 14.5\% & 0.3\% & 18.24\% & -\\
    \bottomrule
    \end{tabular}
    \label{tab:percentage_fgsm_agnews}
\end{table}

\section{Conclusions and Future Work}
\label{sec:conclusions}

In this paper, we mathematically formulated a problem and proposed a framework to implement widely adopted state-of-the-art adversarial attacks in computer vision. To the best of our knowledge, we acknowledge our work as the first to implement such attacks for text classifiers. Then, we designed an encoder-decoder architecture framework such that the input's type is similar to the regenerated text's type, and embedding vectors are well-organized. As a result, our architecture can generate texts with similar meanings to the input text. Our framework can generate adversarial examples that look natural, and the targeted classifier was then misled to predict them as the wrong classes. 

Despite the success of our framework, it still has some limitations. First, training the encoder and decoder for a specific corpus takes a long time. Second, we need to try several values of the perturbation bound ($\epsilon$) to ensure that adversarial examples look the same as the clean ones.

Currently, we plan to experiment with this framework on various datasets and classifiers to show that our attack is vital in many kinds of corpora. All these experiments are scheduled as future work.

\begin{ack}
Part of this work was funded by the National Science Foundation under grants CNS-2136961 and CNS-2210091. We also thank Dr. Gissella Bejarano and Dr. Javier Orduz for their advice in an earlier version of this project.
\end{ack}

\bibliographystyle{plain}
\bibliography{main}

\end{document}